\documentclass[a4paper]{spie}
\usepackage[a4paper, top=2.54cm, bottom=4.94cm, left=1.925cm, right=1.925cm]{geometry}

\usepackage{amsmath,amsfonts,amssymb}
\usepackage{graphicx}
\usepackage{acronym}
\usepackage{xcolor}
\usepackage{tabularray}
\usepackage[normalem]{ulem}
\usepackage{booktabs}
\usepackage{array}
\usepackage{color}
\usepackage[colorlinks=true, allcolors=blue]{hyperref}

\title{SIAMBA: A Pseudo-Siamese Mamba-based Deep Learning Network for Multimodal Flood Change Detection}
\author[a]{Muhammad Fasih Ishfaq}
\author[a]{Giulio Weikmann}
\author[a,*]{Lorenzo Bruzzone}
\authorinfo{*Corresponding author: Lorenzo Bruzzone,
E-mail: \href{mailto:lorenzo.bruzzone@unitn.it}
{lorenzo.bruzzone@unitn.it}}
\affil[a]{Department of Information Engineering and Computer Science, University of Trento, Via Sommarive, 5, I-38123 Trento, Italy}

\begin{document}
\maketitle
\newacro{ALGS}{Adaptive Label Granularity Selection}
\newacro{CCE}{categorical cross-entropy}
\newacro{CCI}{Climate Change Initiative}
\newacro{CEMS}{Copernicus Emergency Management Service}
\newacro{CLC}{CORINE Land Cover}
\newacro{CNN}{Convolutional Neural Network}
\newacro{ConvLSTM}{Convolutional Long Short-Term Memory}
\newacro{CORINE}{COoRdination of INformation on the Environment}
\newacro{CV}{Computer Vision}
\newacro{DL}{Deep Learning}
\newacro{ELU}{Emilia Land Use}
\newacro{EO}{Earth Observation}
\newacro{ESA}{European Space Agency}
\newacro{HRN}{Hierarchical Residual Network}
\newacro{mF1}{mean F1 Score}
\newacro{F1}{F1 Score}
\newacro{FAO}{Food and Agriculture Organization}
\newacro{FCL}{Fully Connected Layer}
\newacro{FM}{Foundation Model}
\newacro{GSB}{Granularity-Specific Block}
\newacro{HAF}{Hierarchy Aware Features}
\newacro{HCL}{Hierarchical Contrastive Learning}
\newacro{HC}{Hierarchical Consensus}
\newacro{HRLC}{High Resolution Land Cover}
\newacro{HXE}{hierarchical cross-entropy}
\newacro{IoU}{Intersection over Union}
\newacro{mIoU}{mean Intersection over Union}
\newacro{MGM}{multi-granularity module}
\newacro{MGC}{Multigranularity Classification}
\newacro{JSD}{Jensen-Shannon Divergence}
\newacro{KLD}{Kullback–Leibler divergence}
\newacro{L2A}{Level-2A}
\newacro{LCCS}{Land Cover Classification System}
\newacro{LC}{Land-Cover}
\newacro{LSE}{log-sum-exp}
\newacro{LSTM}{Long Short-Term Memory}
\newacro{LULC}{Land-Use and Land-Cover}
\newacro{LU}{Land-Use}
\newacro{MGRS}{Military Grid Reference System}
\newacro{ML}{Maximum Likelihood}
\newacro{MS}{multispectral}
\newacro{NLP}{Natural Language Processing}
\newacro{OA}{Overall Accuracy}
\newacro{RF}{Random Forest}
\newacro{RNG}{random number generator}
\newacro{RS}{Remote Sensing}
\newacro{ResNet}{Residual neural network}
\newacro{SAHC}{Semantics-Aware Hierarchical Consensus}
\newacro{SAR}{Synthetic Aperture Radar}
\newacro{SITS}{Satellite Image Time Series}
\newacro{S1}{Sentinel-1}
\newacro{S2}{Sentinel-2}
\newacro{SVM}{Support Vector Machine}
\newacro{SWA}{Stochastic Weight Averaging}
\newacro{TPE}{Tree-structured Parzen Estimator}
\newacro{TS}{time series}
\newacro{ViT}{Vision Transformer}
\newacro{VHR}{Very-High Resolution}

\begin{abstract}
Flood change detection from bi-temporal multispectral and synthetic aperture radar (SAR) images is essential for rapid disaster response. In this context, Sentinel-2 multispectral imagery provides discriminative spectral observations, whereas Sentinel-1 SAR provides weather-independent structural information. Using these modalities jointly requires modelling temporal evidence while accounting for cross-modal heterogeneity. In particular, change detection is complicated by cloud contamination, illumination variability, acquisition geometry and speckle, and heterogeneous land-cover responses. To address these challenges, we propose SIAMBA, a pseudo-siamese Mamba-based network for bi-temporal flood change detection from paired Sentinel-1 and Sentinel-2 observations. SIAMBA uses dedicated modality-specific encoders with dual-pooling squeeze-and-excitation recalibration and selective state-space processing for long-range spatial context. A gated cross-modal change module derives absolute, signed, and relative log-ratio differences at multiple semantic levels, and attention-refined skip connections support progressive multi-scale decoding. Experiments are carried out on OMBRIA benchmark, which includes an independent test set and geographically diverse subsets in France, Albania, Guyana, and Timor. SIAMBA achieves $81.98\%$ IoU, and $93.57\%$ OA on the independent test set, reaching the highest values in the considered comparison. It also provides the highest unweighted average IoU and OA across four regional subsets spatially disjuncted. These results indicate that combining selective state-space contextual modelling with gated multi-scale discrepancy learning effectively improves flood change mapping across the evaluated geographic settings.
\end{abstract}


\section{Introduction}

Rapid, reliable delineation of flooded areas is important for emergency response, impact assessment, and recovery planning. Satellite observations are especially valuable when direct access to affected areas is limited, but flood change mapping is more demanding than static water segmentation. Flooded regions may appear and recede rapidly, generating elongated or fragmented structures, and interact with vegetation, urban surfaces, terrain, and mixed pixels. Moreover, pixel responses depend not only on inundation but also on acquisition conditions and the surrounding spatial context \cite{qin2025rapid,lang2024flood,zhao2024urbansarfloods}.

Multispectral and \ac{SAR} observations offer complementary evidence for this task. Multispectral imagery provides detailed spectral information that can distinguish land cover elements such as water, vegetation, bare soil, and built surfaces, but its reliability may be reduced by cloud cover, shadows, atmospheric effects, and changing illumination. \ac{SAR} operates independently of daylight and penetrates clouds, providing structural and scattering information during adverse weather. However, speckle and variable backscattering from built, vegetated, or rough surfaces can make water-related interpretation ambiguous \cite{drakonakis2022ombrianet,chen2025mdca,lang2024flood}. Multimodal flood studies consequently seek to exploit the strengths of both modalities through approaches ranging from multi-branch flood detection pipelines to multiresolution sensor fusion \cite{qin2025rapid,fawakherji2025flood}. Multimodal approaches alone, however, do not guarantee complementary learned features, as the modalities are governed by different imaging physics and may produce incompatible feature statistics. Effective flood change detection must preserve sensor-specific information while identifying temporal and cross-modal evidence relevant to inundation.

Existing methods address this requirement at different stages of the processing chain. OmbriaNet, which introduced the OMBRIA dataset adopted in this work, performs supervised flood mapping through a multi-branch convolutional architecture operating on bi-temporal \ac{S1} and \ac{S2} observations \cite{drakonakis2022ombrianet}. In heterogeneous change detection, M-UNet establishes cross-sensor comparability through early fusion of concatenated image patches \cite{lv2022unet}, whereas other studies retain sensor-specific branches and learn comparable representations using self-supervised clustering, or contrastive objectives \cite{saha2021self, 9874918}. Other approaches reduce the modality gap more explicitly. GLCD-DA applies image translation before combining edge-sensitive low-level features with global--local CNN--Transformer representations \cite{li2025glcd}, while MMCD uses historical optical observations, temporal memory, and difference enhancement to support change inference when current optical information is unavailable or degraded \cite{zhang2025mmcd}. These formulations demonstrate the importance of sensor-adapted processing, multiscale information, and selective fusion. However, they do not jointly combine supervised bi-temporal \ac{S1}/\ac{S2} flood mapping with efficient long-range contextual modelling and an explicit multilevel representation of cross-modal discrepancies.

Long-range spatial context is particularly relevant because inundated areas may extend over large regions, contain disconnected components, and exhibit irregular boundaries. Convolutional encoders provide strong local feature extraction, but their effective receptive field can be insufficient for relating distant yet spatially connected regions. Global self-attention alleviates this limitation at a computational cost that grows quadratically with the number of spatial tokens. Selective state-space models, including Mamba, instead provide data-dependent long-range propagation with favourable linear sequence scaling \cite{gu2024mamba}. Their suitability for multimodal interaction has been demonstrated by Sigma in RGB-X semantic segmentation \cite{wan2025sigma} and by COMO in multimodal object detection \cite{liu2026como}. These methods establish the potential of state-space modelling for multimodal perception, but they do not address bi-temporal flood change detection or explicitly encode the disagreement between sensor-specific change representations.

To address these limitations, we propose SIAMBA, a pseudo-siamese Mamba-based architecture for supervised flood change detection from paired bi-temporal \ac{S1} and \ac{S2} observations. SIAMBA integrates sensor-specific temporal encoding with explicit cross-modal discrepancy learning. Pre-event and post-event observations are first grouped within each modality and processed by independent encoders, avoiding shared filters between multispectral reflectance and \ac{SAR} backscatter. Selective state-space blocks refine the resulting feature hierarchies with long-range spatial context before cross-modal interaction. At corresponding semantic levels, a learned gate modulates absolute, signed, and relative log-ratio discrepancies between the \ac{S1} and \ac{S2} representations, making complementary and conflicting evidence explicit rather than relying solely on concatenation or implicit attention. The modality-specific and discrepancy features are subsequently combined through attention-refined skip connections and progressive multi-scale decoding to recover both extended inundation patterns and local flood boundaries.

The contributions of this work are as follows:
\begin{itemize}
\item A pseudo-siamese architecture that encodes bi-temporal \ac{S1} and \ac{S2} observations in separate sensor-specific branches before multilevel cross-modal interaction.
\item A gated cross-modal change representation that combines absolute, signed, and relative log-ratio discrepancies using complementary spatial receptive fields.
\item The integration of selective state-space contextual modelling and attention-refined multi-scale decoding to recover broad-scale inundation patterns, fragmented regions, and local boundaries.
\item An evaluation on an independent test set and four geographically distinct flood-event subsets, including comparisons with multimodal change-detection methods and unimodal variants.
\end{itemize}
\section{Proposed Method}

\subsection{Problem formulation}

Let $\mathbf{X}^{a}_{1}$ and $\mathbf{X}^{b}_{1}$ denote pre-event and post-event \ac{S1} observations, and let $\mathbf{X}^{a}_{2}$ and $\mathbf{X}^{b}_{2}$ denote the corresponding \ac{S2} observations. Given the multimodal bi-temporal input $\mathcal{X}=\{\mathbf{X}^{a}_{1},\mathbf{X}^{b}_{1},\mathbf{X}^{a}_{2},\mathbf{X}^{b}_{2}\}$, SIAMBA learns a mapping $f_{\theta}(\mathcal{X})=\hat{\mathbf{Y}}$, where $\hat{\mathbf{Y}}\in[0,1]^{H\times W}$ is a flood change probability map and $\mathbf{Y}\in\{0,1\}^{H\times W}$ is the reference map. The two temporal observations are grouped within each modality, allowing temporal information to be represented before \ac{S1} and \ac{S2} features interact. 

\subsection{Architecture overview}

\begin{figure}
  \centering
  \includegraphics[width=\linewidth]{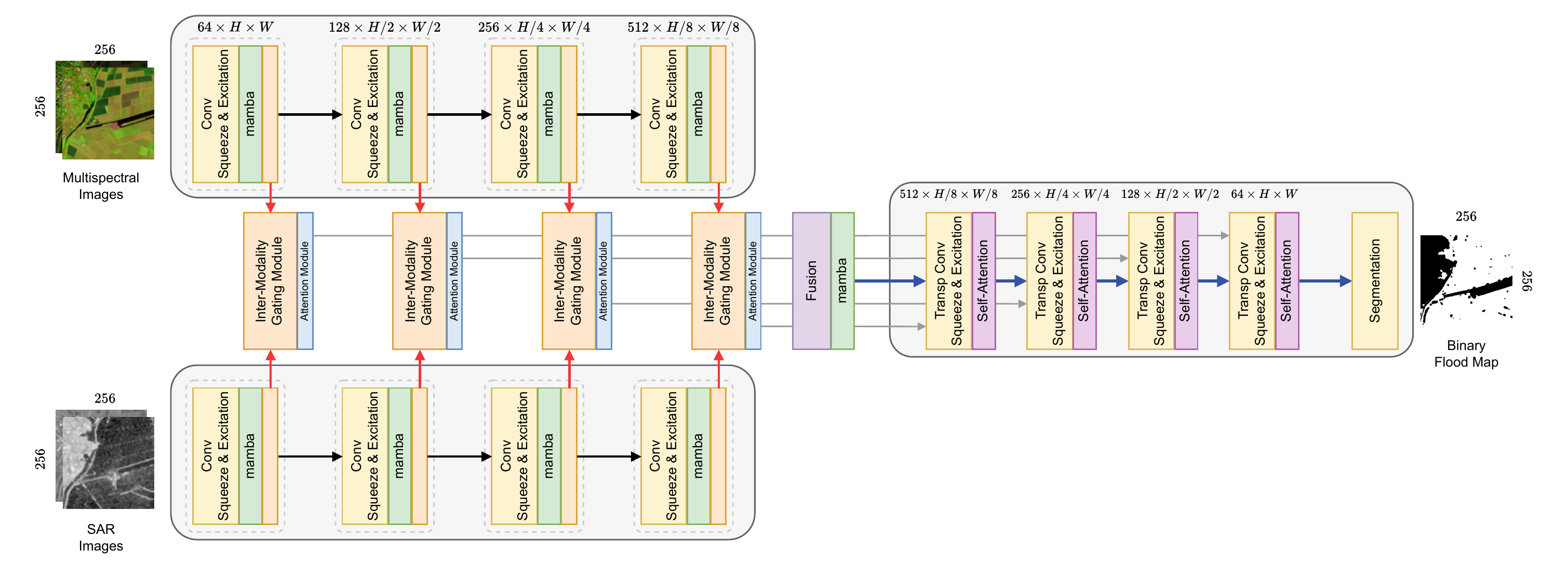}
  \caption{Architecture of the proposed SIAMBA framework. Separate modality-specific encoders extract hierarchical \ac{S1} and \ac{S2} representations, while gated cross-modal change modules support interaction at multiple spatial scales. The fused representation is progressively decoded to obtain the flood change map.}
  \label{fig:architecture}
\end{figure}

Figure~\ref{fig:architecture} summarizes the proposed architecture. The network consists of independent \ac{S1} and \ac{S2} branches, as their inputs represent different physical properties and radiometric distributions. Each branch extracts local spatial features, adaptively recalibrates channel responses, and enriches the representation through spatial state-space processing. Corresponding \ac{S1} and \ac{S2} features then interact through a gated cross-modal change module at several semantic levels. The deepest modality-specific and change features are aggregated into a fused semantic representation. During decoding, the model combines this context with attention-refined modality-specific and change skip features, progressively reconstructing the spatial resolution required for pixel-level flood change prediction.

\subsection{Modality-specific feature extraction}

The two encoder streams process their respective concatenated temporal observations into compatible, independently learned latent spaces. This pseudo-siamese approach avoids imposing identical filters on multispectral reflectance and SAR backscatter, as their sensing physics and radiometric distributions differ substantially. An initial $7\times7$ convolution is followed by four residual stages with 64, 128, 256, and 512 channels. The residual hierarchy preserves local textures and boundaries at shallow levels while increasing the receptive field and semantic abstraction at deeper levels.

Each residual stage includes dual-pooling squeeze-and-excitation recalibration. For a feature map with $C$ channels, global average pooling summarizes the average response of each channel, whereas global maximum pooling retains its strongest activation. The two descriptors are averaged and processed by a bottleneck multilayer perceptron with a reduction factor of $16$ and sigmoid activation, producing channel weights $\mathbf{g}\in[0,1]^C$:
\begin{equation}
\tilde{\mathbf{F}}=\mathbf{g}\odot\mathbf{F}.
\end{equation}
This recalibration combines distributed and peak channel responses, suppresses weak or less relevant channels while preserves modality-specific evidence prior to cross-modal interaction.

Flood extent mapping requires contextual information beyond a small and localized convolutional neighborhood. Large water bodies, disconnected inundated components, and complex shorelines can require information from distant locations to distinguish flood change from locally similar background responses. SIAMBA therefore applies a selective state-space block after each encoder stage, including the deepest level. At encoder level $l$, modality-specific representations $\mathbf{F}_{m}^{l}$ are denoted by:
\begin{equation} \label{eq:encoderfeature}
\mathbf{F}_{m}^{l}=\mathcal{E}_{m}^{l}(\mathbf{X}_{m}), \qquad m\in\{1,2\},
\end{equation}
where $\mathbf{X}_{m}$ denotes the corresponding concatenated bi-temporal input and $\mathcal{E}_{m}^{l}$ is the modality-specific encoder stage. For the modality-specific encoder feature $\mathbf{F}_{m}^{l}$, let $\mathcal{M}^{l}$ denote the corresponding Mamba layer. The operator $\mathcal{R}$ rasterises $\mathbf{F}_{m}^{l}$ in row-major order into a sequence of $H_lW_l$ tokens with embedding dimension $C_l$. This sequence is processed by a single standard Mamba layer with $d=C_l$, after which $\mathcal{R}^{-1}$ restores the original two-dimensional spatial organisation:
\begin{equation}
\widetilde{\mathbf{F}}_{m}^{l}=\mathcal{R}^{-1}\!\left(\mathcal{M}^{l}\!\left(\mathcal{R}(\mathbf{F}_{m}^{l})\right)\right).
\end{equation}
The convolutional hierarchy extracts localized texture and boundary information, while the state-space block propagates contextual information over the spatial-token sequence. Applying this operation at multiple semantic levels allows broader context to influence fine and abstract representations.

\subsection{Cross-modal change representation and feature interaction}

Direct concatenation or summation of \ac{SAR} and multispectral features can mask whether a local response is complementary evidence, modality-specific noise, or cross-modal disagreement. SIAMBA instead forms an explicit cross-modal change representation from corresponding contextual features $\widetilde{\mathbf{F}}_{2}^{l}$ and $\widetilde{\mathbf{F}}_{1}^{l}$ at level $l$. This representation
encodes relations between modality-specific features without replacing their direct extraction and is used alongside them in deep-fusion and skip paths. Although the encoder weights are modality-specific, the two branches are jointly optimised through the discrepancy paths and the common segmentation loss, encouraging task-oriented co-adaptation between corresponding latent channels rather than assuming comparability at the input level.

The module first estimates a data-dependent gate:
\begin{equation}
\mathbf{G}^{l}=\sigma\!\left(\phi_{g}^{l}\!\left([\widetilde{\mathbf{F}}_{2}^{l};\widetilde{\mathbf{F}}_{1}^{l};\max(\widetilde{\mathbf{F}}_{2}^{l},\widetilde{\mathbf{F}}_{1}^{l})]\right)\right),
\label{eq:gate}
\end{equation}
where $[\cdot;\cdot]$ denotes channel concatenation, $\phi_{g}^{l}$ is a learned $1\times1$ projection, and $\sigma$ is the sigmoid function. Considering both feature maps allows the gate to condition on the local evidence of each modality, while the element-wise maximum provides a summary of the stronger local response across the two modalities. The learned projection maps this combined descriptor to a feature-wise gate that regulates subsequent difference information.

\begin{figure}
  \centering
  \includegraphics[width=\linewidth]{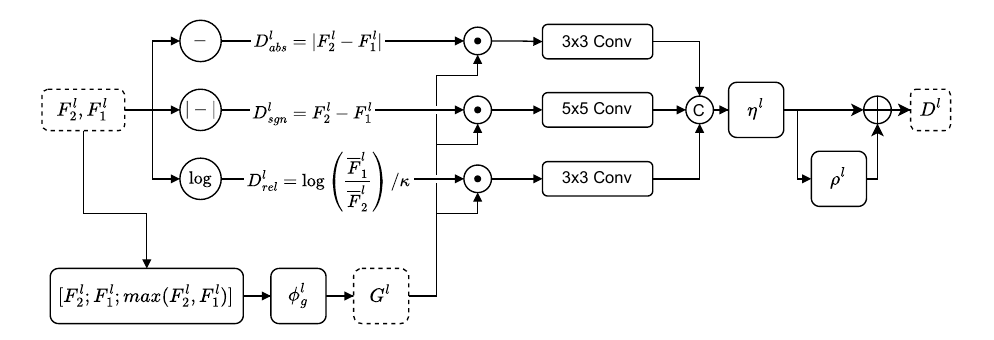}
  \caption{Detailed structure of the gated cross-modal change representation at semantic level $l$. Corresponding Sentinel-1 and Sentinel-2 features estimate a data-dependent gate. Absolute, signed, and relative log-ratio discrepancies are independently modulated and processed with complementary receptive fields before fusion and residual refinement. The resulting representation $\mathbf{D}^{l}$ provides explicit cross-modal evidence to the corresponding skip pathway and deeper multimodal fusion.}
  \label{fig:cross_modal_interaction}
\end{figure}

As illustrated in Figure~\ref{fig:cross_modal_interaction}, the module adopts three complementary discrepancy branches. The three discrepancy quantities considered can be expressed as follows:
\begin{equation}
\mathbf{D}_{\mathrm{abs}}^{l} = |\widetilde{\mathbf{F}}_{2}^{l}-\widetilde{\mathbf{F}}_{1}^{l}|, \qquad \qquad
\mathbf{D}_{\mathrm{sgn}}^{l} = \widetilde{\mathbf{F}}_{2}^{l}-\widetilde{\mathbf{F}}_{1}^{l}, \qquad \qquad
\mathbf{D}_{\mathrm{rel}}^{l} = \frac{1}{\kappa}\log\!\left(\frac{\overline{\mathbf{F}}_{1}^{l}}{\overline{\mathbf{F}}_{2}^{l}}\right),
\end{equation}
where $\mathbf{D}_{\mathrm{abs}}^{l}$ represents the symmetric magnitude of disagreement, $\mathbf{D}_{\mathrm{sgn}}^{l}$ is the signed branch, retaining the direction of the feature relation without assigning a global physical interpretation to the sign, and $\mathbf{D}_{\mathrm{rel}}^{l}$ is the relative log-ratio branch where $\overline{\mathbf{F}}_{m}^{l}$ denotes a clipped positive bounded representation and $\kappa=6$ is a scaling factor. This last branch normalizes the discrepancy by response magnitude and can distinguish proportional variation from a large absolute response.

After extracting the discrepancy quantities, each branch is modulated by the learned gate:
\begin{equation}
\widehat{\mathbf{D}}_{k}^{l}=\mathbf{G}^{l}\odot\mathbf{D}_{k}^{l}, \qquad k\in\{\mathrm{abs},\mathrm{sgn},\mathrm{rel}\},
\end{equation}
and subsequently processed with different receptive fields. The absolute and relative branches use $3\times3$ convolutions for local discrepancy patterns, whereas the signed branch uses a $5\times5$ convolution to capture broader contextual variation. Their responses are concatenated as:
\begin{equation}
\mathbf{Z}^{l}=
\left[
\phi_{3\times3}^{l}(\widehat{\mathbf{D}}_{\mathrm{abs}}^{l});
\phi_{5\times5}^{l}(\widehat{\mathbf{D}}_{\mathrm{sgn}}^{l});
\phi_{3\times3}^{l}(\widehat{\mathbf{D}}_{\mathrm{rel}}^{l})
\right].
\end{equation}
The concatenated representation is fused through a $3\times3$ convolution followed by batch normalization and ReLU activation:
\begin{equation}
\mathbf{U}^{l}=\eta^{l}(\mathbf{Z}^{l}),
\end{equation}
while a subsequent residual transformation refines the fused representation:
\begin{equation}
\mathbf{D}^{l}=\mathbf{U}^{l}+\rho^{l}(\mathbf{U}^{l}),
\label{eq:crossmodal}
\end{equation}
where $\eta^{l}$ and $\rho^{l}$ denote distinct $3\times3$ convolutional blocks with batch normalization and ReLU activation. The resulting feature $\mathbf{D}^{l}$ preserves complementary cross-modal discrepancy information while suppressing unreliable responses through gating. It provides explicit interaction evidence at both shallow and deep semantic resolutions.

\subsection{Multi-scale fusion and decoding}

At the deepest level, the two modality-specific representations and their cross-modal change feature are concatenated, regularized by dropout, and channel-recalibrated to form a fused semantic representation. A decoder with four transposed-convolution stages progressively restores spatial resolution. At each scale, the upsampled feature is combined with the corresponding \ac{S1}, \ac{S2}, and cross-modal change features. A reduced-resolution self-attention block estimates spatial relationships within the skip representation and restores the attended feature to its original scale through bilinear interpolation. This attention-refined fusion emphasizes informative locations while preserving the detailed features needed to recover flood boundaries, fragmented regions, and small changed areas. A final convolutional prediction head maps the decoder output to a single flood change probability map.

\subsection{Optimisation objective}

In flood change mapping, flood change masks are generally imbalanced because changed pixels occupy a smaller fraction of a patch than unchanged pixels. SIAMBA therefore minimizes a weighted binary cross-entropy term together with Dice loss. For $N$ pixels, predicted probabilities $p_i$, binary labels $y_i$, positive-class weight $w$, and numerical constant $\epsilon$, the objective can be defined as follows:
\begin{equation}
\mathcal{L}=-\frac{1}{N}\sum_{i=1}^{N}\left[w y_i\log(p_i)+(1-y_i)\log(1-p_i)\right]+1-\frac{2\sum_i p_i y_i+\epsilon}{\sum_i(p_i+y_i)+\epsilon}.
\end{equation}
The weighted cross-entropy increases the contribution of flooded pixels, while the Dice term promotes spatial overlap between the predicted and reference flood change regions.

\section{Experimental Setup}

\subsection{Dataset and study areas}

To assess the performance of the SIAMBA architecture, we use the OMBRIA dataset, a supervised bitemporal and multimodal flood-mapping benchmark dataset based on co-registered \ac{S1} \ac{SAR} and \ac{S2} multispectral observations with binary reference maps derived from \ac{CEMS} products \cite{drakonakis2022ombrianet}. The dataset contains 3,376 images from 23 flood events acquired between 2017 and 2021. The evaluation includes an independent test set and geographically distinct subsets in France, Albania, Guyana, and Timor. This setup assesses performance across multiple events rather than within a single geographic context.

\subsection{Data preparation and experimental configuration}

Each sample includes co-registered pre-event and post-event \ac{S1} and \ac{S2} image patches with a binary flood change annotation. \ac{S2} is represented by three RGB channels per date and \ac{S1} by one channel (VV) per date, resulting in six \ac{S2} and two \ac{S1} temporal-input channels. Values are normalized to $[0,1]$, and image patches have spatial size $256\times256$ pixels. We followed the official OMBRIA partitioning protocol, using only the provided training set for model training and the predefined test and regional subsets for evaluation. We randomly split the training data into $90\%$ training and $10\%$ validation partitions using a predefined seed for all comparisons. We applied spatial augmentation to all input observations and their corresponding labels, including horizontal flipping, vertical flipping, and rotations by multiples of $90$ degrees. To address class imbalance, we used weighted sampling, assigning higher probabilities to patches with larger flood fractions. 

Table~\ref{tab:configuration} summarizes the training configuration. The model is selected based on the lowest validation loss, and predictions were thresholded at 0.5 during evaluation. Experiments were conducted on a workstation running Ubuntu 22.04.5 LTS, equipped with an AMD Ryzen 9 5950X processor, 128 GB of RAM, and an NVIDIA GeForce RTX 3090 GPU (driver version: 580.142, CUDA version 13.0). The software environment was containerized using the NVIDIA PyTorch 24.06 image and PyTorch 2.4.0. 

\begin{table}
\centering
\caption{Experimental configuration of SIAMBA.}
\label{tab:configuration}
\small
\setlength{\extrarowheight}{1pt} 
\begin{tabular}{ll}
\hline\hline
Setting & Value \\
\hline
Patch size & $256\times256$ pixels \\
\ac{S2} / \ac{S1} temporal channels & 6 / 2 \\
Training epochs / batch size & 50 / 16 \\
Optimizer / initial learning rate & AdamW / $5\times10^{-5}$ \\
Learning-rate schedule & Adaptive reduction on validation-loss plateau (factor 0.5; patience 5 epochs) \\
Loss & weighted BCE + Dice; positive weight 1.5 \\
Augmentation & flips and 90-degree rotations \\
Model selection / decision threshold & lowest validation loss / 0.5 \\
\hline\hline
\end{tabular}
\end{table}

\subsection{Comparison methods and evaluation metrics}

To evaluate the performance of SIAMBA, we compare it against six representative flood change-mapping baselines: (i) OmbriaNet \cite{drakonakis2022ombrianet}, (ii) Sigma \cite{wan2025sigma}, (iii) COMO \cite{liu2026como}, (iv) MMCD \cite{zhang2025mmcd}, (v) GLCD-DA \cite{li2025glcd}, and (vi) M-UNET \cite{lv2022unet}. Furthermore, we benchmark the proposed model against its unimodal variants, SIAMBA \ac{S1}-only and SIAMBA \ac{S2}-only, to assess the impact of multimodal data fusion. All comparison methods were adapted to produce a one-channel binary segmentation output. The same training, validation, augmentation, sampling, and checkpoint-selection protocols were used unless the original method required a task-specific component. Learning rate and weight decay were selected through grid search for all the approaches considered. Positive-class weight has been fixed to $w=1.5$ to prioritize flooded samples. The compared methods span bitemporal multimodal flood mapping, Mamba-based multimodal segmentation, cross-modal interaction, memory-based change detection, global-local attention, multi-scale U-Net decoding, and self-supervised multisensor change detection. 

Performance is evaluated with \ac{IoU} and \ac{OA}. \ac{OA} summarizes all pixels and can remain high when unchanged pixels dominate the prediction. \ac{IoU} provides more direct information about the overlap and quality of the flood change region. 
\section{Results and Discussion}

\subsection{Quantitative comparison}

Table~\ref{tab:results} compares the results obtained by multimodal SIAMBA with those of the seven considered literature methods. On the independent test set, SIAMBA obtains $81.98\%$ \ac{IoU} and $93.57\%$ \ac{OA}. These values exceed the most accurate competing results by $0.76$\% in \ac{IoU} relative to M-UNet ($81.22\%$), and by $0.71$\% in \ac{OA} relative to OmbriaNet ($92.86\%$). 
Across the regional subsets, SIAMBA obtains the highest \ac{IoU} in France, Albania, and Timor, and the highest \ac{OA} in France, Albania, and Guyana. The two exceptions are Guyana \ac{IoU}, where SIAMBA is second to MMCD by $0.56$\% ($76.51\%$ versus $77.07\%$), and Timor \ac{OA}, where it is second to Sigma by a marginal $0.06$\% ($95.62\%$ versus $95.68\%$). SIAMBA therefore ranks first in 6 of the 8 regional dataset--metric combinations. Its regional averages are $75.39\%$ \ac{IoU} and $96.22\%$ \ac{OA}, compared with the strongest competing averages $71.80\%$ \ac{IoU} for M-UNet and $95.09\%$ \ac{OA} for Sigma, corresponding to margins of $3.59$, and $1.13$\%, respectively. This regional pattern suggests a more stable geographic transfer across the evaluated flood events. Across the compared approaches, the results are consistent with the benefit of preserving modality-specific processing before selective interaction, representing discrepancy information explicitly rather than relying only on concatenation or implicit fusion, and combining broad contextual modelling with multiscale boundary recovery.

\begin{table}
\centering
\caption{Quantitative comparison (\%) of SIAMBA with existing flood-mapping and multimodal change-detection methods. The best result for each metric is shown in bold, and the second-best result is underlined.}
\label{tab:results}
\begin{tblr}{
  width = \linewidth,
  colspec = {l l *{7}{X[m,c]}},
  cell{2,4,6,8,10}{1} = {r=2}{},
  cell{2-7,9-10}{9} = {font=\bfseries}, 
  cell{8}{6} = {font=\bfseries}, 
  cell{11}{4} = {font=\bfseries}, 
  vline{3} = {-}{},
  hline{1-2,12} = {-}{},
  hline{1,12} = {2}{-}{},
  hline{4,6,8,10} = {-}{dashed},
}
Dataset & Metric    & OmbriaNet \cite{drakonakis2022ombrianet}        & Sigma \cite{wan2025sigma} & COMO   \cite{liu2026como}       & MMCD  \cite{zhang2025mmcd}        & GLCD-DA\cite{li2025glcd}     & M-UNet \cite{lv2022unet}   & Proposed SIAMBA  \\  
TEST & IoU (\%)    & 74.67         & 77.72 & 81.18         & 79.89         & 78.31         & \uline{81.22}   & 81.98 \\  
     & OA (\%)      & \uline{92.86} & 91.36 & 91.59         & 91.69         & 90.46         & 90.94          & 93.57 \\  

FRANCE & IoU (\%)  & 71.73         & 76.17 & 62.74         & 74.72         & 70.99         & \uline{81.92}   & 85.31  \\  
       & OA (\%)    & \uline{98.48} & 98.13 & 95.40         & 96.98         & 98.38         & 97.48          & 98.86  \\  

ALBANIA & IoU (\%) & 68.31         & 65.78 & 64.02         & 62.40         & 66.90         & \uline{68.54}  & 70.82 \\  
        & OA (\%)   & 89.47         & 90.49 & 90.67         & 90.25         & \uline{91.41} & 90.98         & 92.75 \\  

GUYANA & IoU (\%)  & 70.09         & 73.18 & 72.09         & 77.07         & 69.24         & 75.27           & \uline{76.51}\\  
       & OA (\%)    & 95.19         & 96.07 & 96.29         & \uline{96.66} & 94.73         & 96.37          & 97.64        \\  

TIMOR & IoU (\%)   & \uline{67.06} & 65.70 & 59.15         & 62.83         & 60.38         & 61.46           & 68.92         \\  
      & OA (\%)     & 91.01         & 95.68 & 94.77         & 94.09         & 94.27         & 94.92          & \uline{95.62} \\  
\end{tblr}
\end{table}

\subsection{Contribution of multimodal information}

Table~\ref{tab:unimodal} compares the multimodal model with the \ac{S2}-only and \ac{S1}-only SIAMBA variants. Multimodal SIAMBA exceeds both unimodal variants for all reported datasets and metric. On the independent test set, the multimodal configuration outperforms S2-only SIAMBA by $8.91$\% in \ac{IoU} and $4.31$\% in \ac{OA}, while the corresponding gains over \ac{S1}-only SIAMBA are $15.23$ and $8.37$\%.
The regional-average improvements of multimodal SIAMBA are $18.16$\% in \ac{IoU} and $5.90$\% in \ac{OA} relative to \ac{S2}-only SIAMBA, and $16.21$ and $8.38$\%, respectively, relative to \ac{S1}-only SIAMBA. The relative strength of the individual modalities varies across the evaluated settings. \ac{S1}-only SIAMBA outperforms \ac{S2}-only SIAMBA in France and Guyana, whereas \ac{S2}-only SIAMBA is stronger in Albania, Timor, and the independent test set. This pattern is consistent with complementary modality information across the evaluated settings.

\begin{table}
\centering
\caption{Quantitative comparison (\%) of multimodal SIAMBA with its unimodal variants on the independent test set and regional subsets. The best result for each metric is shown in bold, and the second-best result is underlined.}
\label{tab:unimodal}
\begin{tblr}{
  colspec = {l *{9}{X[c,m]}},
  colsep = 1pt,
  hline{1,3,6} = {-}{},
  hline{1,6} = {2}{-}{},
  vline{4,6,8,10} = {-}{dashed},
  vline{2} = {-}{},
  cell{1}{2,4,6,8,10} = {c=2}{},
  cell{5}{2-11} = {font=\bfseries},
}
& TEST & & FRANCE & & ALBANIA & & GUYANA & & TIMOR & \\
Method          & IoU (\%)      & OA (\%)      & IoU (\%)      & OA (\%)       & IoU (\%)      & OA (\%)        & IoU (\%)      & OA (\%)       & IoU (\%)      & OA (\%) \\
S2-only SIAMBA  & \uline{73.07} & \uline{89.26}& 47.21         & 93.01         & \uline{65.60} & \uline{90.73}  & 57.97         & 89.70         & \uline{58.14} & \uline{87.85} \\
S1-only SIAMBA  & 66.75         & 85.20        & \uline{77.67} & \uline{95.14} & 52.43         & 88.01          & \uline{64.30} & \uline{93.66} & 42.34         & 74.53 \\
Proposed SIAMBA & 81.98         & 93.57        & 85.31         & 98.86         & 70.82         & 92.75          & 76.51         & 97.64         & 68.92         & 95.62 \\
\end{tblr}
\end{table}

\subsection{Qualitative analysis}

\begin{figure*}
  \centering
  \includegraphics[width=\linewidth]{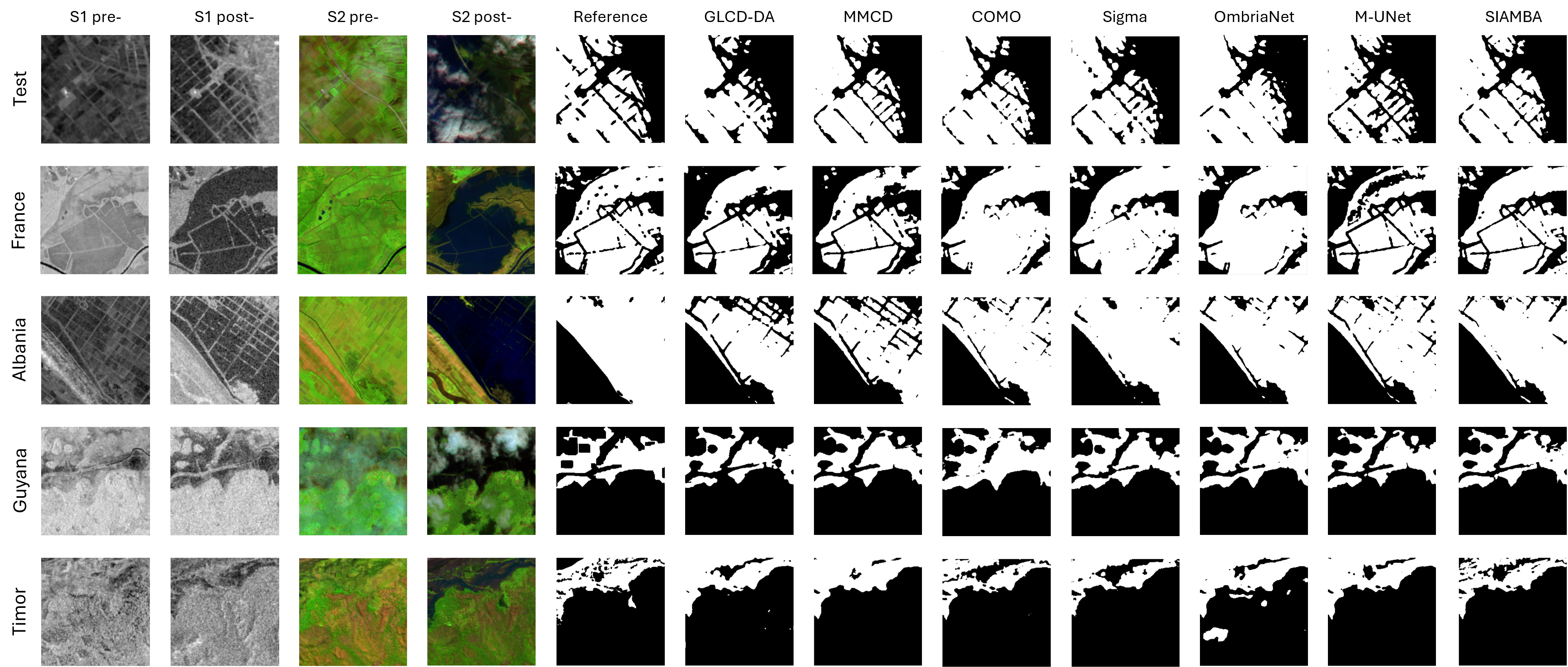}
  \caption{Qualitative comparison of flood change predictions for representative scenes from the independent test set and the France, Albania, Guyana, and Timor subsets. Columns show the pre- and post-event \ac{S1} and \ac{S2} observations, the reference map, and the predictions obtained by GLCD-DA, MMCD, COMO, Sigma, OmbriaNet, and the proposed SIAMBA. Changed pixels are represented in white, whereas unchanged pixels in black.}
  \label{fig:qualitative_results}
\end{figure*}

Figure~\ref{fig:qualitative_results} compares representative scenes from the independent test set and the France, Albania, Guyana, and Timor subsets. Across these geographically diverse cases, SIAMBA generally produces flood change maps with better spatial agreement with the reference annotations, particularly along irregular boundaries and narrow linear structures. In the test and France examples, SIAMBA preserves internal unchanged regions and field boundaries that are omitted by most competing methods. In Albania, it reconstructs the main flood boundary while suppressing noise patterns observed in some predictions. For the Guyana scene, the proposed method performs well overall, although some small elements seem simplified. In Timor, SIAMBA better retains the irregular upper boundary and limits isolated false detections.

Overall, the qualitative results are consistent with the quantitative comparison and indicate that the combination of modality-specific features, gated cross-modal discrepancies, and multi-scale decoding supports both large-scale flood delineation and local boundary recovery. The remaining errors are mainly concentrated in small fragmented regions and locally ambiguous transition areas.

\section{Conclusion}

This paper introduces SIAMBA as a supervised framework for flood change detection from paired bi-temporal \ac{S1} \ac{SAR} and \ac{S2} multispectral observations. The proposed architecture preserves sensor-adapted representations through independent encoder branches, refines them with selective state-space contextual modelling, and explicitly describes their relations through gated absolute, signed, and relative log-ratio discrepancies at multiple semantic levels. These features are integrated with attention-refined skip connections to recover both extended inundation patterns and local spatial details. Experimental validation was conducted on the OMBRIA test set. SIAMBA achieved $81.98\%$ \ac{IoU} and $93.57\%$ \ac{OA}, obtaining the highest values in the considered comparison. It also ranked first in 6 of the 8 regional dataset–-metric combinations and reached regional averages of $75.39\%$ \ac{IoU} and $96.22\%$ \ac{OA}. Moreover, the multimodal configuration consistently outperformed both unimodal variants, whereas the relative performance of \ac{S1}-only and \ac{S2}-only SIAMBA varied across regions, indicating that the two modalities provide complementary information. The quantitative and qualitative findings support the effectiveness of the combined sensor-specific, contextual, and discrepancy-based formulation. Future work will extend the proposed formulation to irregular multi-temporal sequences and investigate the integration of complementary contextual information, such as terrain and hydrological priors, to support reliable near-real-time flood monitoring.

\section*{Acknowledgments}
This work was supported by the Italian Space Agency (ASI) and the Ministry of University and Research (MUR) under Contract 2024-5-E.0 -- CUP n. I53D24000060005, within the ``Space It Up!'' project.

\noindent 
This manuscript is the author-prepared preprint and is not the final published version. The version of record is published in the Proceedings of SPIE and is available at https://doi.org/10.1117/12.3115920.

\bibliographystyle{spiebib}
\bibliography{report,References}
\end{document}